%% file: main_arxiv.tex
\documentclass[letterpaper]{article} 
\usepackage{aaai2026}  
\usepackage{times}  
\usepackage{helvet}  
\usepackage{courier}  
\usepackage[hyphens]{url}  
\usepackage{graphicx} 
\usepackage{natbib}  
\usepackage{caption} 
\usepackage{multirow}
\usepackage{booktabs} 
\usepackage{algorithm}
\usepackage{algorithmic}
\usepackage{amsmath}
\usepackage{amsfonts}
\usepackage{xcolor}
\usepackage[table]{xcolor}
\usepackage{colortbl}
\usepackage{etoc}
\usepackage{fontawesome5}
\usepackage{mdframed}

\usepackage{enumitem}
\setlist[itemize]{leftmargin=1.5em} 

\definecolor{lightgray}{gray}{0.93}

\usepackage{newfloat}
\usepackage{listings}
\DeclareCaptionStyle{ruled}{labelfont=normalfont,labelsep=colon,strut=off} 
\floatstyle{ruled}
\newfloat{listing}{tb}{lst}{}
\floatname{listing}{Listing}
\title{On Sample-Efficient Generalized Planning via Learned Transition Models}

\author{
    Nitin Gupta\equalcontrib,
    Vishal Pallagani\equalcontrib,
    John A. Aydin, 
    Biplav Srivastava
}
\affiliations{
    University of South Carolina\\
    \{niting@email., vishalp@mailbox., jaaydin@email., biplav.s@\}sc.edu
}

\begin{document}

\maketitle

\begin{abstract}

Generalized planning studies the construction of solution strategies that generalize across families of planning problems sharing a common domain model, formally defined by a transition function $\gamma : S \times A \rightarrow S$. Classical approaches achieve such generalization through symbolic abstractions and explicit reasoning over $\gamma$. In contrast, recent Transformer-based planners, such as PlanGPT and Plansformer, largely cast generalized planning as direct action-sequence prediction, bypassing explicit transition modeling. While effective on in-distribution instances, these approaches typically require large datasets and model sizes, and often suffer from state drift in long-horizon settings due to the absence of explicit world-state evolution.
In this work, we formulate generalized planning as a transition-model learning problem, in which a neural model explicitly approximates the successor-state function $\hat{\gamma} \approx \gamma$ and generates plans by rolling out symbolic state trajectories. Instead of predicting actions directly, the model autoregressively predicts intermediate world states, thereby learning the domain dynamics as an implicit world model. To study size-invariant generalization and sample efficiency, we systematically evaluate multiple state representations and neural architectures, including relational graph encodings.
Our results show that learning explicit transition models yields higher out-of-distribution satisficing-plan success than direct action-sequence prediction in multiple domains, while achieving these gains with significantly fewer training instances and smaller models.
\textbf{This is an extended version of a short paper accepted at ICAPS 2026 under the same title.}
\end{abstract}

\begin{links}
    \link{Code}{https://github.com/ai4society/state-centric-gen-planning}
\end{links}

\input{context_files-arxiv/1_introduction}
\input{context_files-arxiv/2_related_work}
\input{context_files-arxiv/3_state_centric_paradigm}
\input{context_files-arxiv/4_experimental_setup}
\input{context_files-arxiv/5_results_analysis}
\input{context_files-arxiv/6_conclusion_future}

\section*{Acknowledgments}

This work is partially supported by NSF Awards \#2454027 and NAIRR250014, and Faculty Award by JP Morgan Research.

\bibliography{aaai2026}

\input{context_files-arxiv/7_appendix}

\end{document}

%% file: context_files-arxiv/1_introduction.tex
\section{Introduction}

\begin{figure*}[htbp]
\centering
\includegraphics[width=\linewidth]{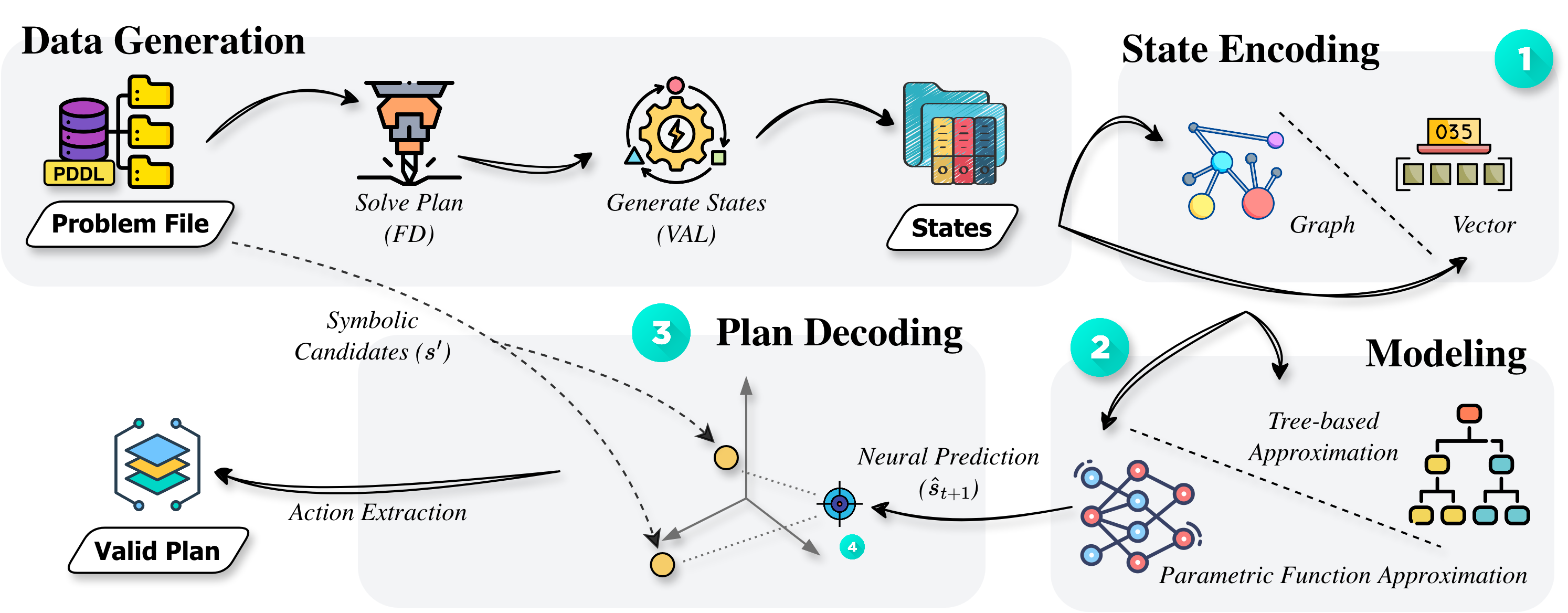}
\caption{
\textbf{State-Centric Generalized Planning Pipeline.}
From a symbolic planning instance $\Pi$, executable plans are generated using a learned transition model.
\textbf{(1) State Encoding:} Symbolic state--goal pairs $(s_t, g)$ are mapped to fixed-dimensional embeddings $\phi(s_t)$ using either WL graph kernels or fixed-size factored vectors.
\textbf{(2) Transition Modeling:} A parametric model (LSTM) or a non-parametric model (XGBoost) learns residual state transitions $\Delta_t$ to predict successor embeddings.
\textbf{(3) Neuro-Symbolic Plan Decoding:} The predicted successor embedding $\hat{\phi}(s_{t+1})$ is matched against all valid symbolic successors $\mathrm{Succ}(s_t)$ induced by $\gamma$, and the nearest valid successor is selected to recover the executable action.
This guarantees symbolic validity while enabling transition-model-based generalization.
}
\label{fig:main_arch}
\end{figure*}

Classical automated planning is defined over a state-transition system $\Sigma = \langle S, A, \gamma \rangle$, where $S$ is the set of states, $A$ the set of actions, and $\gamma : S \times A \rightarrow S$ the transition function. A planning task $\Pi = \langle \Sigma, s_0, g \rangle$ consists of an initial state $s_0 \in S$ and a goal condition $g$, typically represented as a set of literals such that a state $s$ satisfies $g$ iff $g \subseteq s$. A solution is an action sequence $\pi = \langle a_1,\ldots,a_n \rangle$ such that $s_{t+1} = \gamma(s_t,a_t)$ and the final state $s_n$ satisfies $g$. Generalized planning seeks strategies that solve families of such tasks sharing a common $\gamma$. 

Recent learning-based approaches to generalized planning predominantly model the conditional distribution $p({\pi \mid \Pi})$, for example via an autoregressive factorization $p(\pi \mid \Pi) = \prod_{t=1}^{T} p(a_t \mid \Pi, a_{<t})$ or a policy $\pi_\theta(a_t \mid s_t,g)$, and directly predict action sequences from problem descriptions, as in Plansformer \cite{pallagani2022plansformer}, PlanGPT \cite{rossetti2024learning}, and symmetry-aware Transformers \cite{Fritzsche_Gestrin_Seipp_2026}. This action-centric formulation bypasses explicit modeling of $\gamma$: the evolving world state $s_t$ is never directly represented, and long-horizon reasoning relies on implicit correlations between action tokens, leading to state drift in out-of-distribution regimes. 

In this work, we instead model generalized planning as a transition-model learning problem. Rather than predicting the next action, we learn a goal-conditioned neural transition model $\mathcal{T}_\theta$ that predicts the successor state along a plan trajectory, i.e., given the current state $s_t$ and goal $g$ (or problem description $\Pi$), the model outputs a prediction $\hat{s}_{t+1} = \mathcal{T}_\theta(s_t,g)$. Plans are then obtained by rolling out the predicted state trajectory and recovering the corresponding actions via local symbolic search over applicable operators, by matching $\gamma(s_t,a)$ to $\hat{s}_{t+1}$. This formulation enforces explicit world-state evolution, enables successor validation, and constrains learning to respect frame axioms and causal effects. It is consistent with model-based world modeling in reinforcement learning \cite{ha2018world,hafner2019learning}, but applied here to symbolic generalized planning. 

For clarity, we refer to methods that directly predict actions (e.g., modeling $p(\pi \mid \Pi)$ or $\pi_\theta(a_t \mid s_t,g)$) as \emph{action-centric} learning, and to methods that predict successor states via $\mathcal{T}_\theta(s_t,g)$ as \emph{state-centric} learning. This usage is distinct from heuristic-based state-space learning in existing GP taxonomies \cite{chen2025l4p_tutorial} and serves only to distinguish transition-model learning from direct action modeling. 

A central challenge in learning $\mathcal{T}_\theta$ for GP is size invariance: the description of a state in $S$ scales with the number of objects. To address this, we systematically evaluate multiple fixed-dimensional state representations, including fixed-size factored encodings and Weisfeiler--Leman (WL) graph embeddings \cite{chen2024return}. WL embeddings map variable-sized relational states to fixed-length structural feature vectors, enabling compact models such as LSTMs \cite{hochreiter1997long} and XGBoost \cite{chen2016xgboost} to generalize from small to large problem instances. We empirically show that relational WL features are critical for size-invariant and sample-efficient generalization. Our contributions, thus, are

\begin{enumerate}[label=(\roman*)]
    \item a transition-model-based formulation of generalized planning via goal-conditioned successor-state prediction;
    \item a systematic evaluation of state representations for size-invariant and sample-efficient generalization; and
    \item an empirical demonstration that compact models achieve competitive GP performance with orders of magnitude fewer parameters and training instances than Transformer-based planners.
\end{enumerate}

%% file: context_files-arxiv/2_related_work.tex
\section{Related Work}

Learning-based approaches to generalized planning seek policies or models that transfer across problem instances within a domain. Early neural GP work introduced relational inductive biases to enable size generalization, including Action Schema Networks \cite{toyer2018action} and graph-based deep RL for Blocksworld \cite{rivlin2020generalized}, as well as finite-state policy representations \cite{staahlberg2022learning}. More recent work formulates GP as sequence prediction: Plansformer \cite{pallagani2022plansformer} and PlanGPT \cite{rossetti2024learning} train Transformer architectures to directly generate action sequences, while symmetry-aware Transformers \cite{Fritzsche_Gestrin_Seipp_2026} introduce architectural and contrastive constraints to improve permutation invariance. However, such action-centric models do not explicitly learn transition dynamics and often exhibit state drift under distributional shift. In parallel, graph-based state representations have been extensively studied for learning heuristics and value functions in planning, including STRIPS-HGN \cite{shen2020learning}, GOOSE \cite{chen2023goose}, and domain-independent graph transformations \cite{chen2024graph}. Recent work shows that Weisfeiler--Leman (WL) graph kernels combined with lightweight regressors can match or exceed GNN performance at far lower cost \cite{chen2024return,chen2025weisfeiler,hao2025effective}. Hybrid neuro-symbolic systems integrate learned components with symbolic solvers or validators, including LLM+P \cite{liu2023llm+}, symbolic validation for PlanGPT \cite{rossetti2024enhancing}, LLM-Modulo \cite{kambhampati2024position}, and SayCan for robotics \cite{ahn2022can}. Separately, model-based reinforcement learning demonstrates the benefits of learning explicit world models for planning \cite{ha2018world,hafner2019learning}. In contrast to prior action-sequence and heuristic-centric neural planners, our work adopts a transition-prediction formulation of generalized planning with size-invariant state representations, enabling sample-efficient and robust out-of-distribution generalization using compact models.
Extended discussion is in Appendix~\ref{sec:extended_related}.

%% file: context_files-arxiv/3_state_centric_paradigm.tex
\section{Transition-Model-Based Generalized Planning}
\label{sec:formulation}

\begin{algorithm}[b]
\caption{\textit{State-Centric} GP with Plan Decoding}
\label{alg:state-centric-planning}
\begin{algorithmic}[1]
\REQUIRE Initial state $s_0$, goal $g$, operators $\mathcal{A}$, learned model $f_\theta$, embedding $\phi$
\ENSURE Valid plan $\pi$
\STATE $t \leftarrow 0$, $\pi \leftarrow \langle \rangle$, $s_t \leftarrow s_0$
\WHILE {$g \not\subseteq s_t$}
    \STATE \hspace{0.8em} $\mathbf{v}_t \leftarrow \phi(s_t) + f_\theta(\phi(s_t), \phi(g))$
    \STATE \hspace{0.8em} $\mathrm{Succ}(s_t) \leftarrow {\{\gamma(s_t,a) \mid a \in \mathcal{A},\; a \text{ applicable in } s_t\}}$
    \STATE \hspace{0.8em} $s_{t+1} \leftarrow \arg\min_{s' \in \mathrm{Succ}(s_t)} \|\phi(s') - \mathbf{v}_t\|_2$
    \STATE \hspace{0.8em} $a_t \leftarrow$ unique $a$ such that $\gamma(s_t,a) = s_{t+1}$
    \STATE \hspace{0.8em} $\pi. \text{append}(a_t)$, $s_t \leftarrow s_{t+1}$, $t \leftarrow t+1$
\ENDWHILE
\RETURN $\pi$
\end{algorithmic}
\end{algorithm}

Figure~\ref{fig:main_arch} illustrates the complete pipeline of our approach, consisting of symbolic data generation, size-invariant state encoding, transition-model learning, and plan decoding via symbolic verification.

\paragraph{Generalized Planning Setup.}
A planning instance is $\Pi=\langle \mathcal{O},\mathcal{P},\mathcal{A},s_0,g\rangle$, where $\mathcal{O}$ is a finite object set, $\mathcal{P}$ a predicate vocabulary, $\mathcal{A}$ a set of operators, $s_0\subseteq\mathcal{P}(\mathcal{O})$ the initial state, and $g\subseteq\mathcal{P}(\mathcal{O})$ the goal condition. Operators induce a deterministic transition function $\gamma:S\times\mathcal{A}\rightarrow S$. A plan $\pi=\langle a_1,\ldots,a_T\rangle$ satisfies $s_{t+1}=\gamma(s_t,a_t)$ and $g\subseteq s_T$. Generalized planning seeks a single parameterized model trained on instances from $\mathcal{D}_{\text{train}}$ (small $|\mathcal{O}|$) that generalizes to $\mathcal{D}_{\text{test}}$ with much larger $|\mathcal{O}|$.

\paragraph{Size-Invariant State Representation.}
Each state-goal pair $(s,g)$ is encoded as a relational instance graph $G_{s,g}$ and embedded using $k$ iterations of WL color refinement. Node color histograms yield a fixed-dimensional embedding $\phi(s,g)\in\mathbb{R}^D$, where $D$ depends only on the domain and is independent of $|\mathcal{O}|$. 
We overload notation and write $\phi(s)$ and $\phi(g)$ for the state and goal components, respectively.
The resulting representation is permutation-invariant, size-invariant, and as expressive as 1-WL message-passing GNNs while enabling lightweight downstream models.

\paragraph{State-Centric Transition-Model Learning.}
Rather than learning a policy $\pi_\theta(a_t\mid s_t,g)$, we learn a neural transition model $f_\theta:\mathbb{R}^D\times\mathbb{R}^D\rightarrow\mathbb{R}^D$ that predicts state updates in embedding space.
We denote this embedding-space transition model as $f_\theta$; the conceptual model $\mathcal{T}_\theta$ from the introduction is realized as

\begin{equation*}
    \mathcal{T}_\theta(s_t,g) \approx \phi^{-1}(\phi(s_t) + f_\theta(\phi(s_t),\phi(g)))
\end{equation*}

where the inverse is approximated via nearest-neighbor decoding.
To exploit the sparsity of STRIPS-style transitions, where most predicates remain unchanged, we adopt a residual formulation:

\begin{equation*}
    \hat{\phi}(s_{t+1})=\phi(s_t)+f_\theta(\phi(s_t),\phi(g))
\end{equation*}

where $f_\theta$ predicts a delta vector $\Delta_t$. This explicitly encodes frame axioms and improves sample efficiency, particularly for non-sequential models. We train by minimizing the squared error over expert trajectories:

\begin{equation*}
    {\mathcal{L}=\sum_{t} \|\hat{\phi}(s_{t+1})-\phi(s_{t+1})\|_2^2}.
\end{equation*}

\paragraph{Plan Decoding via Neuro-Symbolic Verification.}
At test time, the true symbolic state $s_t$ is maintained throughout execution. Given $s_t$, the transition model produces a target embedding $\mathbf{v}_t=\phi(s_t)+f_\theta(\phi(s_t),\phi(g))$. Using the symbolic operators, we enumerate all valid successors

\begin{equation*}
    \mathrm{Succ}(s_t)=\{\gamma(s_t,a)\mid a\in\mathcal{A},\; a \text{ applicable in } s_t\}
\end{equation*}

and select the successor whose embedding is closest to the neural prediction:

\begin{equation*}
    s_{t+1}=\arg\min_{s'\in\mathrm{Succ}(s_t)} \|\phi(s')-\mathbf{v}_t\|_2.
\end{equation*}

The executed action is the unique $a$ satisfying $\gamma(s_t,a)=s_{t+1}$, which is well-defined under deterministic operators. This decoding step guarantees symbolic validity at every timestep and performs online correction of neural prediction errors. The procedure terminates when $g\subseteq s_t$. The full planning algorithm is summarized in Algorithm~\ref{alg:state-centric-planning}.

%% file: context_files-arxiv/4_experimental_setup.tex
\section{Experimental Setup}
\label{sec:experiments}

We evaluate whether learning an explicit transition model enables sample-efficient and size-invariant generalized planning. Our experiments assess: 
\begin{enumerate}[label=(\roman*)]
    \item extrapolative out-of-distribution (OOD) generalization from small to large instances,
    \item whether parametric function approximation is necessary for learning transition dynamics, and
    \item the impact of size-invariant state representations.
\end{enumerate}

\paragraph{Domains and Data.}
We evaluate on 4 IPC benchmark domains: \textit{Blocksworld}, \textit{Gripper}, \textit{Logistics}, and \textit{VisitAll}. Problem instances are sourced from the Symmetry-Aware Transformer repository, following the same data splits for fair comparison. Symbolic plans are generated using Fast Downward~\cite{helmert2006fast} with the landmark-cut heuristic, and complete state trajectories are reconstructed using VAL \cite{howey2004val}.
Data is partitioned into four splits by object count: \emph{Training} (small instances, e.g., 4--7 blocks), \emph{Validation} (similar or slightly larger sizes, held out), \emph{Interpolation} (unseen configurations within the training size range), and \emph{Extrapolation} (strictly larger than any training instance, e.g., 9--17 blocks). Extrapolation is the primary evaluation axis for size-invariant generalized planning.
Full dataset statistics are reported in Appendix section~\ref{sec:dataset_details}.

\paragraph{State Representations.}
We compare WL graph embeddings~\cite{chen2024graph}, which are permutation- and size-invariant, against Fixed-Size Factored (FSF) encodings.
FSF encodings represent states as fixed-dimensional vectors with pre-assigned object slots, deliberately omitting the relational structure of WL to isolate the contribution of invariant representations to OOD generalization \cite{boutilier2000stochastic,guestrin2003efficient}. Details about both representations are in Appendix sections~\ref{sec:wl_details} and~\ref{sec:fsf_details}.

\paragraph{Transition Models.}
We evaluate two transition-model classes: a parametric neural model (two-layer LSTM) and a tree-based, nonparametric regressor (XGBoost). The LSTM tests whether sequential memory is necessary for trajectory-level transition dynamics, while XGBoost tests whether a local approximation of the transition kernel suffices. This comparison isolates the role of temporal memory. Both models are trained in state-prediction and delta-prediction modes, as defined in previous sections. The entire technical pipeline and hyperparameters are reported in Appendix section~\ref{sec:impl_pipeline}.

\paragraph{Baselines.}
We compare against published results from Symmetry-Aware Transformers (SymT), which include results on PlanGPT, including applicability-filtered and regrounded variants.
We further run inference with Plansformer on our test instances using its publicly released checkpoint; it was trained on Blocksworld but not on Gripper, Logistics, or VisitAll, so its zero-shot cross-domain performance is expected to be limited.
PlanGPT results are taken directly from \citet{Fritzsche_Gestrin_Seipp_2026}, who train and evaluate PlanGPT on the same data splits and counts as SymT; the weak interpolation/extrapolation performance reflects the difficulty of learning generalized policies from small training sets via action-centric sequence prediction.
As a symbolic upper bound, we include Fast Downward with A* and the landmark-cut heuristic. While alternative configurations such as LAMA~\cite{richter2010lama} may improve coverage on extrapolation instances at varying runtimes, we retain A*+LM-cut as a fixed reference; the neural baselines serve primarily to compare learning paradigms rather than to benchmark against optimized classical planners.

\paragraph{Inference and Metrics.}
At test time, plans are generated using the neuro-symbolic decoding procedure in
Algorithm~\ref{alg:state-centric-planning} with beam width 3 and a dynamic horizon cap of
$T_{\max} = \max(100,\; 10 \cdot |\mathcal{O}|)$, where $|\mathcal{O}|$ is the number of
objects in the test instance (see Appendix~\ref{sec:dynamic_horizon}).
Performance is measured by \emph{satisficing success rate}, i.e., the fraction of instances for which a generated plan is valid under the transition model and reaches a goal state within the horizon limit, as verified by VAL. We additionally report changes in satisficing success across successive rollouts (seeds) to quantify stability under repeated decoding.

\begin{table*}[htbp]
\centering
\caption{
    Coverage rates (\%) across all configurations. Values are reported as $Mean_{\pm Std}$. 
    Best result per row (over all non-FD configurations) is \textbf{bolded}, second best is \underline{underlined}. 
    Light gray columns denote our state-centric implementations.
    FD = Fast Downward (60s timeout). $^*$Results from \citet{Fritzsche_Gestrin_Seipp_2026}. 
}
\label{tab:full_results}

\newcommand{\val}[2]{#1\ensuremath{_{\scriptscriptstyle \pm #2}}}
\newcommand{\best}[2]{\textbf{#1}\ensuremath{_{\scriptscriptstyle \pm \mathbf{#2}}}}
\newcommand{\secb}[2]{\underline{#1}\ensuremath{_{\scriptscriptstyle \pm \underline{#2}}}}
\newcommand{\zer}[0]{0.00\ensuremath{_{\scriptscriptstyle \pm 0.00}}}

\setlength{\tabcolsep}{2.5pt}

\resizebox{\textwidth}{!}{%
\begin{tabular}{ll|c|c|ccc|c|ccc|
>{\columncolor{lightgray}}c
>{\columncolor{lightgray}}c|
>{\columncolor{lightgray}}c
>{\columncolor{lightgray}}c|
>{\columncolor{lightgray}}c
>{\columncolor{lightgray}}c|
>{\columncolor{lightgray}}c
>{\columncolor{lightgray}}c}
\toprule
& & \textbf{FD} & \textbf{Plansf.} & \multicolumn{3}{c|}{\textbf{PlanGPT}$^*$} & \textbf{SymT$^\text{E}$}$^*$ & \multicolumn{3}{c|}{\textbf{SymT$^\text{ED}$}$^*$} & \multicolumn{2}{c|}{\cellcolor{lightgray}\textbf{WL-LSTM}} & \multicolumn{2}{c|}{\cellcolor{lightgray}\textbf{WL-XGB}} & \multicolumn{2}{c|}{\cellcolor{lightgray}\textbf{FSF-LSTM}} & \multicolumn{2}{c}{\cellcolor{lightgray}\textbf{FSF-XGB}} \\
\textbf{Domain} & \textbf{Split} & & & greedy & appl. & regr. & greedy & greedy & appl. & regr. & state & delta & state & delta & state & delta & state & delta \\
\midrule
\multirow{3}{*}{\textit{Blocks}}
& Val.    & 1.00 & \best{1.00}{0.00} & \zer & \zer & \zer & \best{1.00}{0.00} & \best{1.00}{0.00} & \best{1.00}{0.00} & \zer & \val{0.44}{0.16} & \secb{0.67}{0.00} & \best{1.00}{0.00} & \secb{0.67}{0.00} & \zer & \zer & \zer & \zer \\
& Interp. & 1.00 & \best{1.00}{0.00} & \val{0.56}{0.16} & \val{0.56}{0.16} & \zer & \best{1.00}{0.00} & \best{1.00}{0.00} & \best{1.00}{0.00} & \best{1.00}{0.00} & \val{0.67}{0.27} & \secb{0.89}{0.16} & \best{1.00}{0.00} & \best{1.00}{0.00} & \val{0.11}{0.16} & \zer & \zer & \zer \\
& Extrap. & 0.60 & \val{0.10}{0.00} & \zer & \zer & \zer & \val{0.05}{0.07} & \val{0.07}{0.02} & \val{0.13}{0.05} & \zer & \val{0.10}{0.07} & \val{0.15}{0.07} & \secb{0.25}{0.00} & \best{0.50}{0.00} & \zer & \zer & \zer & \zer \\
\midrule
\multirow{3}{*}{\textit{Gripper}}
& Val.    & 1.00 & \zer & \zer & \zer & \zer & \best{1.00}{0.00} & \val{0.17}{0.24} & \best{1.00}{0.00} & \best{1.00}{0.00} & \best{1.00}{0.00} & \secb{0.67}{0.47} & \zer & \zer & \val{0.17}{0.24} & \val{0.50}{0.00} & \zer & \zer \\
& Interp. & 1.00 & \zer & \zer & \val{0.44}{0.16} & \zer & \secb{0.89}{0.16} & \val{0.67}{0.00} & \best{1.00}{0.00} & \best{1.00}{0.00} & \best{1.00}{0.00} & \val{0.67}{0.47} & \zer & \zer & \val{0.44}{0.31} & \val{0.22}{0.31} & \val{0.67}{0.00} & \zer \\
& Extrap. & 0.13 & \zer & \zer & \zer & \zer & \val{0.02}{0.03} & \zer & \val{0.15}{0.06} & \best{0.79}{0.16} & \secb{0.42}{0.16} & \val{0.25}{0.31} & \zer & \zer & \val{0.04}{0.03} & \zer & \zer & \zer \\
\midrule
\multirow{3}{*}{\textit{VisitAll}}
& Val.    & 1.00 & \zer & \zer & \val{0.14}{0.12} & \zer & \best{1.00}{0.00} & \val{0.33}{0.09} & \val{0.93}{0.04} & \secb{0.99}{0.02} & \best{1.00}{0.00} & \val{0.79}{0.29} & \best{1.00}{0.00} & \best{1.00}{0.00} & \val{0.08}{0.07} & \val{0.53}{0.02} & \val{0.12}{0.00} & \val{0.92}{0.00} \\
& Interp. & 1.00 & \zer & \val{0.05}{0.04} & \val{0.67}{0.18} & \val{0.41}{0.22} & \best{1.00}{0.00} & \val{0.87}{0.01} & \secb{0.99}{0.01} & \best{1.00}{0.00} & \best{1.00}{0.00} & \secb{0.99}{0.01} & \best{1.00}{0.00} & \best{1.00}{0.00} & \val{0.57}{0.02} & \val{0.39}{0.03} & \val{0.86}{0.00} & \val{0.95}{0.00} \\
& Extrap. & 0.50 & \zer & \zer & \val{0.02}{0.02} & \zer & \val{0.42}{0.11} & \zer & \val{0.15}{0.05} & \val{0.64}{0.12} & \secb{0.72}{0.13} & \val{0.62}{0.39} & \val{0.15}{0.00} & \best{1.00}{0.00} & \zer & \val{0.07}{0.01} & \val{0.01}{0.00} & \val{0.16}{0.00} \\
\midrule
\multirow{3}{*}{\textit{Logistics}}
& Val.    & 1.00 & \zer & \zer & \secb{0.08}{0.12} & \zer & \zer & \zer & \zer & \zer & \best{0.25}{0.35} & \val{0.08}{0.12} & \zer & \zer & \zer & \zer & \zer & \zer \\
& Interp. & 1.00 & \zer & \val{0.07}{0.05} & \secb{0.44}{0.09} & \val{0.19}{0.14} & \val{0.11}{0.00} & \val{0.22}{0.31} & \val{0.26}{0.29} & \val{0.22}{0.31} & \best{0.85}{0.14} & \val{0.33}{0.16} & \val{0.11}{0.00} & \val{0.11}{0.00} & \val{0.19}{0.10} & \val{0.11}{0.00} & \val{0.44}{0.00} & \val{0.11}{0.00} \\
& Extrap. & 0.26 & \zer & \zer & \zer & \zer & \zer & \zer & \zer & \zer & \zer & \zer & \zer & \zer & \zer & \zer & \zer & \zer \\
\bottomrule
\end{tabular}%
}
\end{table*}

%% file: context_files-arxiv/5_results_analysis.tex
\section{Results and Analysis}

Table~\ref{tab:full_results} reports satisficing-plan success rates across all domains, splits, representations, model classes, and prediction modes. The primary empirical finding is that explicit transition-model learning combined with size-invariant relational representations yields stronger or matching extrapolation than action-centric sequence prediction in domains with locally factored domains, while remaining insufficient for the Logistics benchmark under strict size extrapolation.

\paragraph{Comparison with action-centric planners.}
Under strict extrapolation, Plansformer and all PlanGPT variants achieve $0.00$ success across all four domains.
Plansformer further exhibits $0.00$ on non-\textit{Blocksworld} domains since they are not present in its training data.
SymT attains non-zero extrapolation in \textit{Blocksworld} ($0.13$), \textit{Gripper} ($0.79$), and \textit{VisitAll} ($0.64$), but fails in \textit{Logistics}.

The best state-centric models exceed SymT in \textit{Blocksworld} (WL-XGB delta $0.50$ vs.\ $0.13$) and \textit{VisitAll} ($1.00$ vs.\ $0.64$), while SymT remains superior in \textit{Gripper} extrapolation ($0.79$ vs.\ $0.42$).
Notably, these gains are obtained using compact transition models trained on unaugmented state trajectories: our LSTM requires only ${\sim}$1.1M--2.1M parameters and XGBoost ${\sim}$128K--819K tree nodes (depending on the domain), compared to ${\sim}$25--35M (SymT), ${\sim}$125M (PlanGPT), and ${\sim}$220M (Plansformer).
Furthermore, SymT augments training data via symmetry-based state-space expansion, whereas our models are trained on the original small training sets without any augmentation (e.g., 9 instances in Blocksworld). This indicates that, under appropriate relational abstractions, explicit transition learning can match or exceed extrapolation at orders-of-magnitude lower model and data cost, suggesting that learning domain physics provides a stronger inductive bias for generalization than architectural scale or data augmentation alone.
A detailed breakdown of parameter counts and training data requirements is provided in Appendix section~\ref{sec:efficiency_analysis}.

\paragraph{Effect of size-invariant relational representations.}
Across all domains, FSF-based encodings yield negligible extrapolation performance: \textit{Blocksworld} ($0.00$), \textit{Gripper} ($0.00$), \textit{VisitAll} ($\leq 0.13$), and \textit{Logistics} ($0.00$). In contrast, WL-based models achieve strictly positive extrapolation in three domains.
In \textit{Blocksworld}, WL-XGB (delta) reaches $0.50$ compared to $0.00$ for all FSF variants. In \textit{VisitAll}, WL-XGB (delta) reaches $1.00$ while FSF-XGB (delta) reaches only $0.16$.
This establishes that extrapolation beyond the training object bound requires a permutation- and size-invariant abstraction $\phi:S\rightarrow\mathbb{R}^D$. Fixed-slot encodings restrict the hypothesis class to a bounded object universe and therefore fail when $|\mathcal{O}|_{\text{test}} > |\mathcal{O}|_{\text{train}}$.

\paragraph{Effect of residual transition modeling.}
For tree-based models, residual parameterization consistently improves extrapolation.
In \textit{Blocksworld}, WL-XGB improves from $0.25$ (state) to $0.50$ (delta).
In \textit{VisitAll}, it improves from $0.15$ to $1.00$.
This behavior is consistent with STRIPS transition semantics,
\(
\gamma(s,a)= (s \setminus \text{Del}(a)) \cup \text{Add}(a),
\)
which induces sparse state differences. The delta formulation constrains learning to the subspace of changed fluents, reducing regression variance for non-parametric models. For LSTM, the effect is domain dependent: delta improves \textit{Blocksworld} extrapolation ($0.03 \rightarrow 0.15$) but degrades \textit{Gripper} ($0.25 \rightarrow 0.17$), indicating interaction between residual bias and recurrent state memory.

\paragraph{Sequential versus non-sequential transition learning.}
Comparing WL-LSTM and WL-XGB isolates the role of temporal memory.

In \textit{Gripper} extrapolation, WL-LSTM (state) attains $0.42$ while both XGB variants remain at $0.00$, indicating that under the chosen abstraction the induced transition kernel
$P(\phi(s_{t+1}) \mid \phi(s_t), \phi(g))$
is not well-approximated by a purely local regressor.
In contrast, in \textit{Blocksworld} and \textit{VisitAll}, WL-XGB (delta) outperforms WL-LSTM:
\textit{Blocksworld} ($0.50$ vs.\ $0.15$) and \textit{VisitAll} ($1.00$ vs.\ $0.72$),
indicating that the Markovian assumption suffices under relational abstraction in these domains.

\paragraph{Limitations under hierarchical causal coupling.}
All learned models, including all state-centric variants, achieve $0.00$ extrapolation in \textit{Logistics}.
Even Fast Downward degrades from $1.00$ (validation) to $0.26$ (extrapolation) under a 60-second timeout, reflecting the exponential state-space growth of extrapolation instances.
The \textit{Logistics} domain exhibits deep multi-layer causal coupling across object types and transport modalities, which is not preserved under local successor matching.
This identifies a concrete structural limitation of one-step neural transition prediction under strict size extrapolation for this domain.

\paragraph{Summary of empirical findings.}
The results support three data-grounded conclusions: 
\begin{enumerate}[label=(\roman*)]
    \item size-invariant relational representations are necessary for extrapolation beyond training object bounds;
    \item residual (delta) modeling significantly improves non-parametric transition learning in sparse STRIPS domains; and
    \item the necessity of sequential memory in transition learning is domain dependent.
\end{enumerate}

Transition-model learning alone, however, remains insufficient for hierarchical domains under strict extrapolation.
An extended experimental analysis is presented in Appendix section~\ref{sec:experimental_figures}.

%% file: context_files-arxiv/6_conclusion_future.tex
\section{Conclusion and Future Work}

We presented a state-centric formulation of generalized planning in which models learn to predict successor states rather than action sequences.
When combined with size- and permutation-invariant relational embeddings, this approach enables compact transition models (${\sim}$1--2M parameters, no data augmentation) to achieve strong extrapolation performance in locally factored domains, matching or exceeding Transformer baselines (${\sim}$25--220M parameters) that rely on orders of magnitude more data and parameters.
Empirically, our results show that explicit transition modeling provides a stronger inductive bias for extrapolation than architectural scale alone, though limitations remain in domains with hierarchical and long-range dependencies. The neuro-symbolic decoding interface further improves robustness by enforcing symbolic validity at every planning step. Future work will target hierarchical and long-range dependency domains, where one-step state prediction fails under strict extrapolation. We will extend the state-centric framework to multi-step or abstract transitions while preserving symbolic verification.

%% file: context_files-arxiv/7_appendix.tex
\clearpage
\appendix
\setcounter{secnumdepth}{2}
\renewcommand{\thesection}{\Alph{section}}

\section*{Appendix}

This appendix provides supplementary material to support reproducibility and extended analysis. It is organized as follows:

\addcontentsline{toc}{section}{Appendix Contents}

\vspace{0.5em}
\noindent\ref{sec:notation}. Notation and Terminology \dotfill \pageref{sec:notation}

\vspace{0.5em}
\noindent\ref{sec:extended_related}. Extended Related Work \dotfill \pageref{sec:extended_related}

\vspace{0.5em}
\noindent\ref{sec:experimental_figures}. Extended Experimental Analysis \dotfill \pageref{sec:experimental_figures}

\vspace{0.5em}
\noindent\ref{sec:dataset_details}. Dataset Details and Statistics \dotfill \pageref{sec:dataset_details}

\vspace{0.5em}
\noindent\ref{sec:wl_details}. Weisfeiler--Leman Graph Embedding Details \dotfill \pageref{sec:wl_details}

\vspace{0.5em}
\noindent\ref{sec:fsf_details}. Fixed-Size Factored Encoding Details \dotfill \pageref{sec:fsf_details}

\vspace{0.5em}
\noindent\ref{sec:impl_pipeline}. Technical Implementation Pipeline \dotfill \pageref{sec:impl_pipeline}

\section{Notation and Terminology}
\label{sec:notation}

We summarize all symbols and technical terms used throughout the paper for clarity and reproducibility in Table~\ref{tab:notation}.

\begin{table*}[htbp]
\centering
\caption{Summary of notation used throughout the paper.}
\label{tab:notation}
\renewcommand{\arraystretch}{1.15}
\resizebox{\textwidth}{!}{%
\begin{tabular}{llp{9cm}}
\toprule
\textbf{Category} & \textbf{Symbol} & \textbf{Meaning} \\
\midrule

\multirow{9}{*}{\textbf{Planning}}
& $\Sigma = \langle S, A, \gamma \rangle$ & State-transition system (set of states $S$, actions $A$, and transition function $\gamma$) \\
& $S$ & Set of all symbolic world states \\
& $A, \mathcal{A}$ & Set of grounded actions (operators) \\
& $\gamma : S \times \mathcal{A} \rightarrow S$ & Deterministic transition function \\
& $\mathcal{O}$ & Finite object set in a planning instance \\
& $\mathcal{P}$ & Predicate vocabulary \\
& $\mathcal{P}(\mathcal{O})$ & Set of all grounded predicates over $\mathcal{O}$ \\
& $s \subseteq \mathcal{P}(\mathcal{O})$ & A symbolic state \\
& $\Pi = \langle \Sigma, s_0, G \rangle$; $\Pi = \langle \mathcal{O}, \mathcal{P}, \mathcal{A}, s_0, g \rangle$ & Planning task (global and instance-level formulations) \\
\midrule

\multirow{5}{*}{\textbf{Goals \& Successors}}
& $s_0$ & Initial state of a planning instance \\
& $G \subseteq S$ & Set of goal states in the state-transition formulation \\
& $g \subseteq \mathcal{P}(\mathcal{O})$ & Goal condition in the propositional formulation \\
& $\mathrm{Succ}(s)$ & Set of valid symbolic successors of state $s$ under $\gamma$ \\
& $s_{t+1} = \gamma(s_t,a_t)$ & Successor state obtained by applying action $a_t$ in $s_t$ \\
\midrule

\multirow{9}{*}{\textbf{Learning \& Transitions}}
& $\mathcal{D}_{\text{train}}, \mathcal{D}_{\text{test}}$ & Training and test distributions over planning instances \\
& $\mathcal{T}_\theta$ & Learned goal-conditioned transition model \\
& $\hat{s}_{t+1}$ & Predicted successor state from $\mathcal{T}_\theta$ \\
& $\phi : S \rightarrow \mathbb{R}^D$ & Fixed-dimensional state embedding \\
& $\phi(g)$ & Embedded representation of the goal condition \\
& $f_\theta$ & Delta-prediction function in the residual transition parameterization \\
& $\Delta_t$ & Residual transition update at step $t$ \\
& $\hat{\phi}(s_{t+1})$ & Predicted next-state embedding, e.g., $\hat{\phi}(s_{t+1}) = \phi(s_t) + \Delta_t$ \\
& $\mathbf{v}_t$ & Target embedding used for decoding at step $t$ \\
\midrule

\multirow{4}{*}{\textbf{Decoding}}
& $s_{t+1} = \displaystyle \arg\min_{s' \in \mathrm{Succ}(s_t)} \|\phi(s') - \mathbf{v}_t\|_2$ & Neuro-symbolic successor selection rule \\
& $a_t = \gamma^{-1}(s_t \rightarrow s_{t+1})$ & Action inducing the selected symbolic transition \\
& $T$ & Plan horizon (length of $\pi$) \\
& $\pi = \langle a_1,\dots,a_T\rangle$ & Plan as an action sequence \\
\midrule

\multirow{5}{*}{\textbf{Representations}}
& WL & Weisfeiler--Leman (graph-kernel) state embedding \\
& FSF & Fixed-Size Factored state encoding \\
& ILG & Instance Learning Graph (relational encoding of $(s,g)$) \\
& $G_{s,g}$ & Relational instance graph encoding the pair $(s,g)$ \\
& $k$ & Number of WL refinement iterations \\
& $D$ & Dimensionality of the embedding $\phi(s) \in \mathbb{R}^D$ \\
\midrule

\multirow{6}{*}{\textbf{Models}}
& LSTM & Parametric recurrent transition model \\
& XGBoost (XGB) & Tree-based non-parametric regressor \\
& State mode & Direct prediction of $\phi(s_{t+1})$ \\
& Delta mode & Prediction of residual $\Delta_t$ with reconstruction $\phi(s_t) + \Delta_t$ \\
& $\theta$ & Trainable parameters of the learned models \\
& $\hat{\gamma}$ & Learned approximation of the symbolic transition function $\gamma$ \\
\midrule

\multirow{4}{*}{\textbf{Evaluation}}
& Validation / Interpolation / Extrapolation & Size-based dataset splits by object count \\
& Satisficing plan & Any valid plan whose execution reaches a goal state \\
& FD & Fast Downward planner (A* with LM-cut heuristic, 60s timeout) \\
& Coverage / Success rate & Fraction of test instances for which a satisficing plan is found \\
\bottomrule
\end{tabular}%
}
\end{table*}

\section{Extended Related Work}
\label{sec:extended_related}

This section expands on the related work discussion in the main paper, providing additional context and citations for each research direction relevant to our work.

\subsection{Learning for Generalized Planning}

Generalized Planning (GP) seeks solution strategies that transfer across problem instances within a domain, rather than solving each instance independently \cite{segovia2021generalized}. Early neural approaches introduced relational inductive biases to enable such transfer. Action Schema Networks (ASNs) \cite{toyer2018action} exploited the lifted action structure of planning domains by constructing neural architectures that mirror action schemas, enabling policies to generalize beyond training instances. \citet{rivlin2020generalized} combined graph neural networks with deep reinforcement learning to learn Blocksworld policies that scale to instances orders of magnitude larger than those seen during training. These works established that an appropriate relational structure is critical for extrapolation in learned planners. More recent approaches have explored diverse architectural choices for GP. \citet{staahlberg2022learning} investigated learning general policies represented as finite-state controllers, providing theoretical grounding for the expressivity required for GP.

\subsection{Symmetry and Invariance in Neural Planning}

A fundamental challenge in neural planning is handling the symmetries inherent in planning problems: permuting object names should not change the solution structure. This has motivated architectures with explicit invariance properties. Graph Neural Networks (GNNs) naturally handle permutation invariance through message-passing operations \cite{battaglia2018relational}.

The Symmetry-Aware Transformer (SymT) framework \cite{Fritzsche_Gestrin_Seipp_2026} addresses symmetry in sequence-to-sequence planning. It introduces three innovations: compositional tokenization (separating objects and predicates), removal of positional encodings (preventing memorization of absolute positions), and contrastive learning objectives (encouraging mapping of symmetric states to similar representations). SymT achieves state-of-the-art performance on several IPC domains through extensive data augmentation (state-space expansion).

Our work differs from SymT in two fundamental respects. First, we adopt a \textit{state-centric} rather than \textit{action-centric} objective: predicting successor states rather than next actions. This forces the model to learn transition dynamics explicitly, providing natural grounding that architectural modifications alone cannot guarantee. Second, we achieve competitive generalization \emph{without} massive data augmentation, relying instead on size-invariant representations (WL embeddings) that provide structural inductive bias.

\subsection{Graph-Based State Representations}

To handle variable-sized object sets, recent planners use graph-based state representations where objects are nodes and predicates define edges or features. Hypergraph networks such as STRIPS-HGN \cite{shen2020learning} learned domain-independent heuristics directly from planning graphs. \citet{chen2023goose} proposed GOOSE, a GNN-based heuristic learner that significantly outperformed STRIPS-HGN and generalized to much larger problem instances. Subsequent work \cite{chen2024return} showed that classical Weisfeiler-Leman (WL) graph kernels \cite{weisfeiler1968reduction,shervashidze2011weisfeiler} with simple regression models can match or exceed GNN performance at orders of magnitude lower computational cost. This finding motivates our use of WL embeddings: they provide the expressivity of 1-WL message-passing GNNs while enabling deployment with lightweight models.

Recent extensions have further improved graph-based representations for planning. \citet{chen2024graph} introduced domain-independent graph transformations that enhance generalization, while \citet{chen2025weisfeiler} provided theoretical analysis connecting WL expressivity to planning-specific structures. \citet{hao2025effective} demonstrated effective combinations of WL features with neural architectures for learning domain-independent heuristics.

\subsection{Action-Centric Sequence Models for Planning}

Another line of work formulates plan generation as a sequence prediction task. Plansformer \cite{pallagani2022plansformer} fine-tuned a Transformer on symbolic plan traces to generate action sequences with high validity on classical planning benchmarks, substantially outperforming zero-shot language models. More recently, PlanGPT \cite{rossetti2024learning} trained GPT-style architectures directly on planning data to learn general planning policies. While effective in-distribution, such autoregressive models often suffer from state drift and logical inconsistency on longer horizons or out-of-distribution problems due to the lack of explicit state tracking. Concurrent work by \citet{shlomi2025transition} explores LLM-based transition prediction from $(s,a)$ pairs, but targets single-instance prediction rather than size-invariant generalized planning. 

\paragraph{Hybrid LLM--Symbolic Planning.}
To mitigate these limitations, several hybrid frameworks integrate LLMs with symbolic planners or validators. LLM+P \cite{liu2023llm+} uses an LLM to generate PDDL formulations that are solved by a classical planner, while recent PlanGPT variants incorporate symbolic validation to enforce action applicability \cite{rossetti2024enhancing}. \citet{kambhampati2024position} formalized this paradigm as \emph{LLM-Modulo}, advocating tight coupling between LLMs and model-based solvers. In robotics, SayCan \cite{ahn2022can} combines language models with learned affordance models to iteratively select feasible actions for long-horizon task execution. 

\paragraph{Model-Based Learning and World Models.}
In reinforcement learning, learning explicit transition dynamics has proven beneficial for planning and generalization. World Models \cite{ha2018world} and subsequent model-based RL methods demonstrate that agents can learn compact environment simulators and plan via internal rollouts. However, most neural planners for classical planning remain action-centric or heuristic-driven and do not explicitly learn the symbolic transition dynamics.
DreamerV3~\cite{hafner2023mastering} extends world-model learning to diverse RL domains, motivating explicit dynamics modeling.

\paragraph{Positioning of Our Work.}
In contrast to prior action-sequence and heuristic-centric approaches, our work adopts a \textit{state-centric, model-based learning} paradigm in which the planner is trained to predict state transitions directly. This enables explicit state grounding at every step, mitigates state drift, and yields significantly improved robustness on out-of-distribution instances using compact models. To the best of our knowledge, this transition-prediction formulation with size-invariant representations has not been previously explored for generalized neural planning.

\section{Extended Experimental Analysis}
\label{sec:experimental_figures}

This section provides additional experimental analysis, including performance trends across problem sizes and per-problem breakdowns.

\subsection{Performance Across Data Sets}

\begin{figure*}[!htbp]
    \centering
    \includegraphics[width=0.85\linewidth]{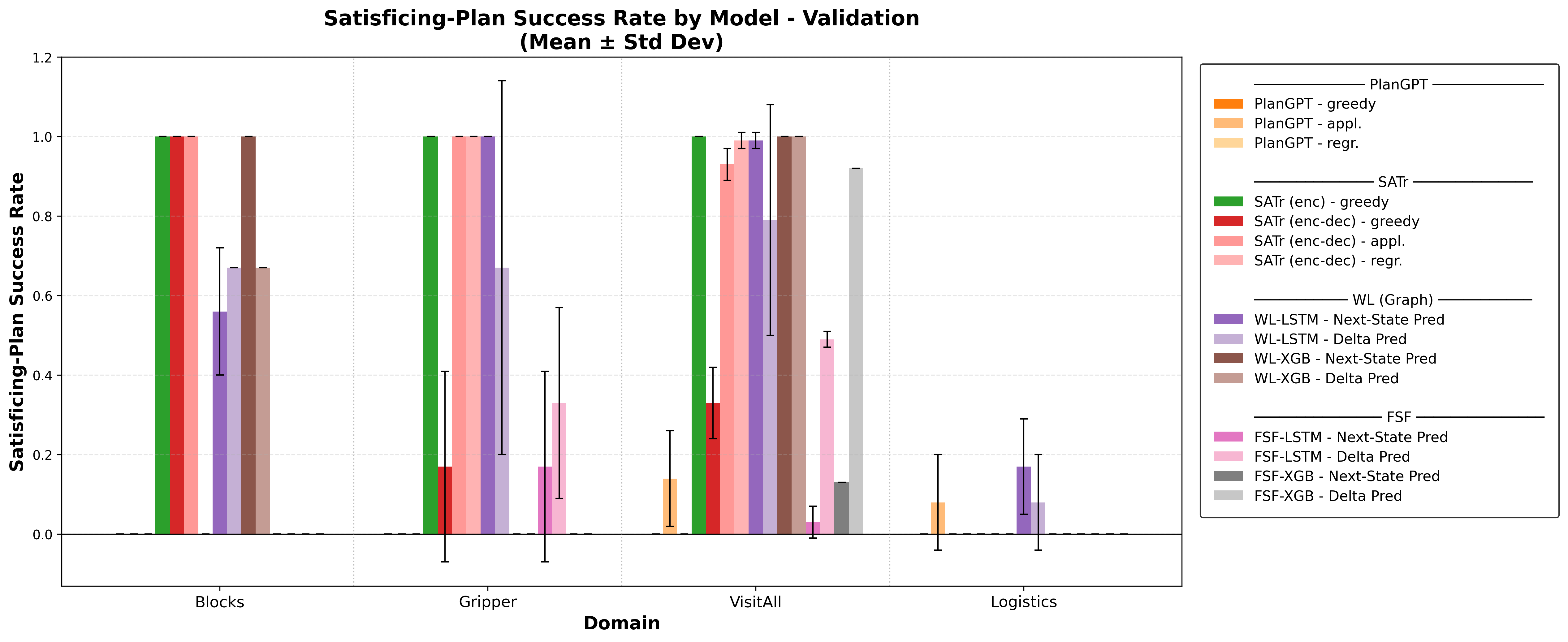}
    \caption{Satisficing-plan success rates on the validation split across all domains, comparing PlanGPT, SymT, WL-based, and FSF baselines.}
    \label{fig:val_bar}
\end{figure*}

\begin{figure*}[!htbp]
    \centering
    \includegraphics[width=0.85\linewidth]{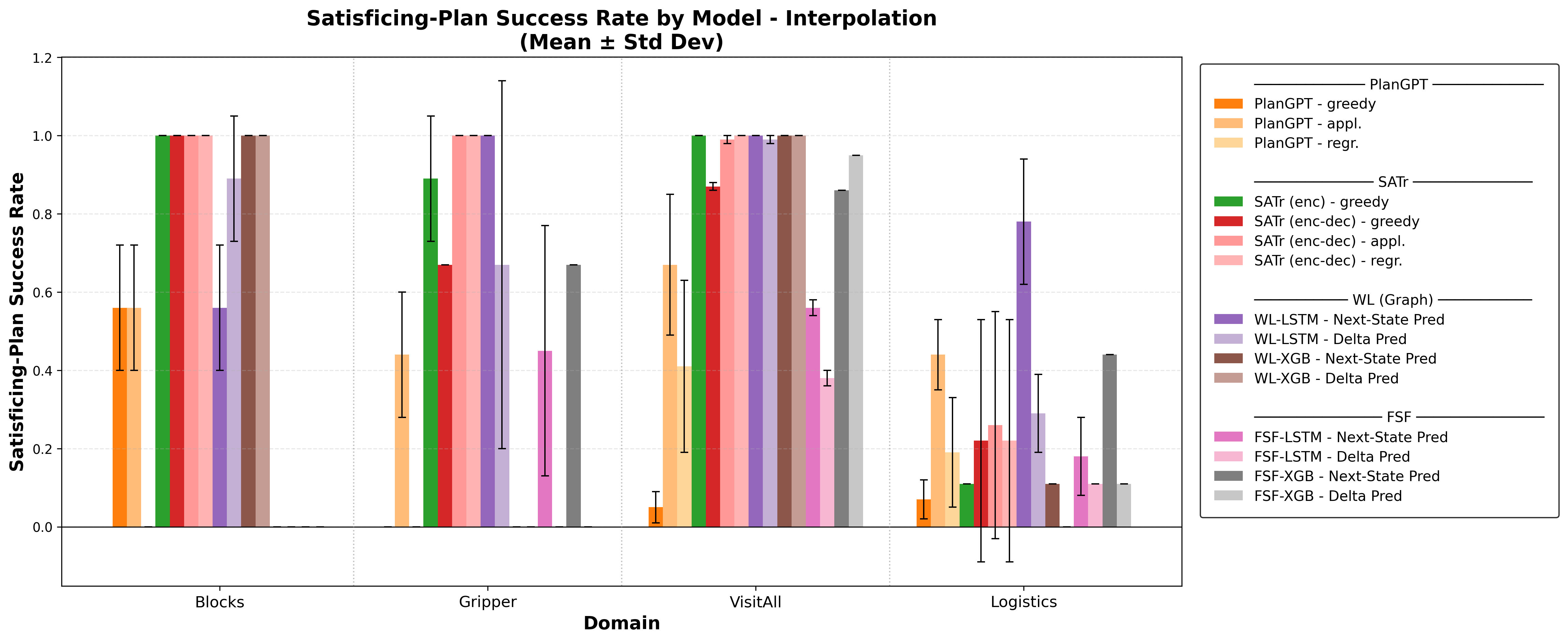}
    \caption{Satisficing-plan success rates on the interpolation split, comparing PlanGPT, SymT, WL-based, and FSF baselines for generalization to in-distribution problem instances.}
    \label{fig:interp_bar}
\end{figure*}

\begin{figure*}[!htbp]
    \centering
    \includegraphics[width=0.85\linewidth]{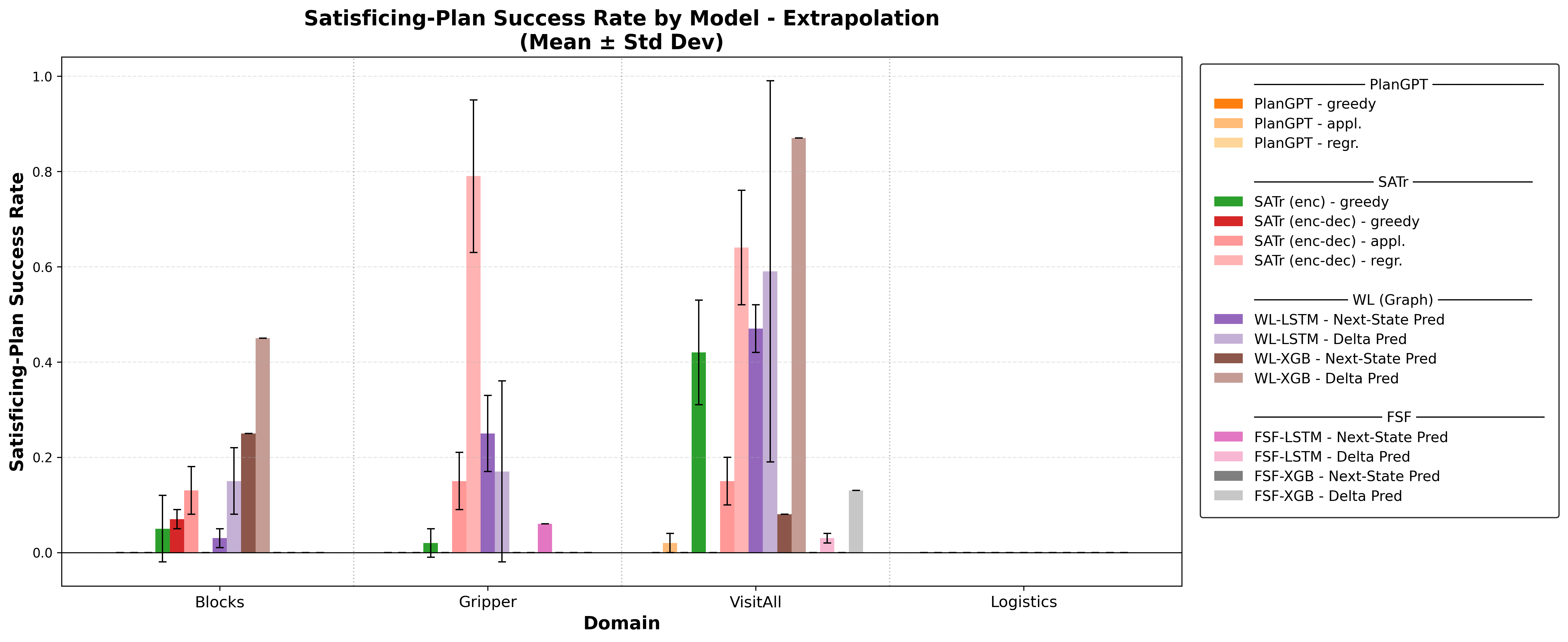}
    \caption{Satisficing-plan success rates on the extrapolation split, comparing PlanGPT, SymT, WL-based, and FSF baselines for generalization in out-of-distribution problem instances.}
    \label{fig:extrap_bar}
\end{figure*}

Figures~\ref{fig:val_bar}--\ref{fig:extrap_bar} show satisficing-plan success rate across different data splits, with plots showing the validation, interpolation, and extrapolation splits for each domain. These figures visualize the data presented in Table 1, providing a more intuitive view of performance trends across problem sizes and model configurations.

\begin{table}[h]
\centering
\small 
\setlength{\tabcolsep}{4pt} 
\caption{State representation dimensions ($D$) and model complexity.}
\label{tab:model_complexity}
\begin{tabular}{lcccc}
\toprule
\textbf{Domain} & \shortstack{\textbf{FSF Dim.}\\\textbf{($D$)}} & \shortstack{\textbf{WL Dim.}\\\textbf{($D$)}} & \shortstack{\textbf{WL-LSTM}\\\textbf{Params}} & \shortstack{\textbf{WL-XGBoost}\\\textbf{Nodes}} \\
\midrule
\textit{Blocksworld} & 17 & 172 & 1,253,292 & 335,154 \\
\textit{Gripper}     & 46 & 302 & 1,552,942 & 128,310 \\
\textit{VisitAll}    & 121 & 138 & 1,174,922 & 679,060 \\
\textit{Logistics}   & 37 & 552 & 2,129,192 & 819,320 \\
\bottomrule
\end{tabular}
\end{table}

\begin{table}[htbp]
\centering
\caption{Model size comparison across all methods. SymT estimates are derived from architectural specifications in \citet{Fritzsche_Gestrin_Seipp_2026} (12-layer shared-weight encoder/decoder, hidden size 768); exact counts depend on domain-specific vocabulary embeddings, which are not reported in the SymT work.}
\label{tab:model_size}
\resizebox{\columnwidth}{!}{
\begin{tabular}{lccc}
\toprule
\textbf{Method} & \textbf{Parameters} & \textbf{Architecture} & \textbf{Ratio} \\
\midrule
Plansformer & $\sim$220M & CodeT5 & $\sim$230$\times$ \\
PlanGPT & $\sim$125M & GPT-2 & $\sim$130$\times$ \\
SymT$^\text{ED}$ & $\sim$25--35M & Transformer & $\sim$30$\times$ \\
\midrule
\textbf{Ours (LSTM)} & $\sim$1.1--2.1M & 2-layer LSTM & 1$\times$ \\
\textbf{Ours (XGB)} & $\sim$340--820K nodes & Gradient boosting & --- \\
\bottomrule
\end{tabular}
}
\end{table}

\subsection{Model and Data Efficiency Analysis}
\label{sec:efficiency_analysis}

A core claim of our state-centric formulation is that explicit transition-model learning enables both sample efficiency and compact model sizes compared to direct action-sequence prediction.

\paragraph{Model Architecture and Complexity.}
Table~\ref{tab:model_complexity} details the exact state representation dimensions ($D$) and the resulting parameter counts for our learned transition models across all domains. 
For the Fixed-Size Factored (FSF) encodings, $D$ is strictly bounded by the maximum number of objects present in the extrapolation split, plus one global slot (e.g., for the robot's location). For the Weisfeiler--Leman (WL) graph encodings, $D$ is determined dynamically by the vocabulary size of unique graph hashes collected during the WL color refinement process on the training set. 

Despite the variations in $D$ across domains, the LSTM models consistently require roughly 1.1--2.1M parameters (using a hidden dimension of 256 and 2 recurrent layers). Similarly, the XGBoost models utilize approximately 128--819K tree nodes (trained with 1000 estimators and a maximum depth of 8). This confirms that our state-centric approach achieves strong OOD generalization using models that are 1 to 2 orders of magnitude smaller than the 25--220M parameters typical of action-centric Transformers.

\paragraph{Training Data Requirements.}
Unlike Transformer-based baselines that often require massive datasets or symmetry-based state-space augmentation, our models are trained strictly on the original, unaugmented small training sets.
For instance, our models achieve $50\%$ strict extrapolation success in \textit{Blocksworld} and $100\%$ in \textit{VisitAll} after training on only 9 and 207 small problem instances, respectively.
In contrast, action-centric models like SymT require extensive data augmentation to achieve even 13\% extrapolation, while PlanGPT and Plansformer fail entirely despite being trained on thousands of instances. This demonstrates that learning the domain physics provides a much stronger inductive bias for generalization than scaling up data and model size.
Table~\ref{tab:model_size} summarizes model sizes across all methods.

\section{Dataset Details and Statistics}
\label{sec:dataset_details}

\begin{table*}[htbp]
\centering
\caption{Dataset statistics. Each cell shows \emph{count (complexity values)} for the corresponding split. Complexity is measured in blocks, balls, goals, or cells depending on the domain. Training uses small instances; extrapolation tests generalization to significantly larger problems.}
\label{tab:dataset_stats}
\resizebox{\textwidth}{!}{%
\begin{tabular}{lcccc}
\toprule
\textbf{Domain} & \textbf{Train} & \textbf{Validation} & \textbf{Interpolation} & \textbf{Extrapolation} \\
\midrule
Blocksworld (blocks) &
9 (4, 6, 7) &
3 (8) &
3 (5) &
20 (9--17) \\

Gripper (balls) &
4 (2, 4, 6, 8) &
2 (9, 10) &
3 (3, 5, 7) &
16 (12--42, even) \\

Logistics (goals) &
12 (1, 3, 5) &
4 (6) &
9 (2, 4) &
18 (7--15) \\

VisitAll (cells) &
207 (1, 3, 4, 6, 10, 11, 12, 14, 16) &
24 (18, 20) &
37 (2, 5, 8, 9, 15) &
219 (24--121) \\
\bottomrule
\end{tabular}%
}
\end{table*}

\subsection{Domain Descriptions}

We evaluate on four standard IPC benchmark domains that admit a classical STRIPS-style representation. Each domain is specified by a tuple $\langle \mathcal{O}, \mathcal{P}, \mathcal{A} \rangle$, where $\mathcal{O}$ is the object set, $\mathcal{P}$ is the predicate vocabulary, and $\mathcal{A}$ is the operator set inducing a deterministic transition function $\gamma(s,a) = s \setminus \text{Del}(a) \cup \text{Add}(a)$. Detailed dataset statistics for each split are reported in Table~\ref{tab:dataset_stats}.

\paragraph{Blocksworld.}
Blocksworld is a canonical manipulation domain defined over a set of blocks $\mathcal{O} = \{b_1,\ldots,b_n\}$. The predicate set includes $\mathcal{P} = \{\textit{on}(x,y), \textit{ontable}(x), \textit{clear}(x), \textit{holding}(x), \textit{handempty}\}$. The operator set $\mathcal{A}$ consists of \textsc{pickup}, \textsc{putdown}, \textsc{stack}, and \textsc{unstack}. A state $s \subseteq \mathcal{P}(\mathcal{O})$ encodes a partial order over blocks. The branching factor grows quadratically in the number of clear blocks due to all admissible \textsc{stack}/\textsc{unstack} combinations. Transitions exhibit sparse add--delete structure, making this domain well-suited for residual state prediction.

\paragraph{Gripper.}
Gripper models a robot with two grippers transporting balls between rooms. The object set factorizes as $\mathcal{O} = \mathcal{B} \cup \mathcal{R} \cup \{\textit{robot}\}$, where $\mathcal{B}$ are balls and $\mathcal{R}$ are rooms. Predicates include $\textit{at}(x,r)$, $\textit{free}(g)$, $\textit{carry}(x,g)$, and $\textit{at-robot}(r)$. Operators include \textsc{move}, \textsc{pick}, and \textsc{drop}. Although single-step transitions remain sparse, Gripper induces longer causal chains in which object transport must be synchronized with robot motion, creating trajectory-level dependencies not captured by purely local transitions.

\paragraph{Logistics.}
Logistics is a multi-modal transportation domain with object types $\mathcal{O} = \mathcal{P} \cup \mathcal{T} \cup \mathcal{A} \cup \mathcal{C}$ (packages, trucks, airplanes, cities). Predicates include $\textit{at}(x,l)$, $\textit{in}(x,v)$, and $\textit{at-vehicle}(v,l)$. Operators include \textsc{load}, \textsc{unload}, \textsc{drive}, and \textsc{fly}. Unlike Blocksworld and VisitAll, Logistics induces long-range dependencies across heterogeneous object types and transport layers. Valid plans require coordination across multiple abstraction levels, resulting in deep coupling between distant subgoals. This property severely limits the effectiveness of one-step transition prediction under strict size extrapolation.

\paragraph{VisitAll.}
VisitAll is a grid navigation domain defined over a lattice of cells $\mathcal{O} = \{(i,j)\}$. Predicates include $\textit{at}(r,c)$ and $\textit{visited}(c)$. Operators correspond to unit grid motions that update both $\textit{at}$ and $\textit{visited}$. The goal is $g = \bigwedge_{c \in \mathcal{O}} \textit{visited}(c)$. Although each transition affects only a small number of fluents, the goal conjunct grows linearly with $|\mathcal{O}|$, and the resulting state space grows exponentially. This domain isolates the effect of goal scaling under otherwise simple local dynamics.

\paragraph{Relevance to State-Centric Modeling.}
All four domains admit deterministic STRIPS semantics with sparse add--delete operators. This ensures that successor states admit a decomposition
\[
s_{t+1} = s_t \setminus \text{Del}(a_t) \cup \text{Add}(a_t),
\]
which directly motivates our residual transition formulation
\(
\hat{\phi}(s_{t+1}) = \phi(s_t) + \Delta_t.
\)
The domains thus provide a controlled testbed for evaluating whether learned neural approximations $\hat{\gamma}$ can preserve symbolic transition structure under strict size extrapolation.

\section{Weisfeiler--Leman Graph Embedding Details}
\label{sec:wl_details}

This section provides a comprehensive description of the Weisfeiler--Leman (WL) graph embedding procedure used in our state-centric planning framework.

\subsection{Background: The WL Algorithm}

The Weisfeiler--Leman (WL) algorithm \cite{weisfeiler1968reduction} is an iterative color refinement procedure for graphs. Given a graph $G = (V, E)$ with initial node colors $c^{(0)}: V \rightarrow \Sigma$, the algorithm iteratively refines colors by aggregating neighborhood information:
\begin{multline*}
    c^{(k+1)}(v) = \\
    \textsc{Hash}\left(c^{(k)}(v), \{\!\{(c^{(k)}(u), \ell(u,v)) \mid u \in \mathcal{N}(v)\}\!\}\right)
\end{multline*}
where $\mathcal{N}(v)$ denotes the neighbors of $v$, $\ell(u,v)$ is the edge label, and $\{\!\{\cdot\}\!\}$ denotes a multiset. After $k$ iterations, the algorithm produces a multiset of colors $\mathcal{C}^{(k)} = \bigcup_{i=0}^{k} \{\!\{c^{(i)}(v) \mid v \in V\}\!\}$.

\subsection{Instance Learning Graph Construction}

We use the Instance Learning Graph (ILG) representation \cite{chen2024return} to encode planning states. Given a planning instance $\Pi = \langle \mathcal{O}, \mathcal{P}, \mathcal{A}, s, g \rangle$, the ILG $G_{s,g} = (V, E, \mathbf{F}_{\text{cat}}, \mathbf{L})$ is constructed as follows:

\paragraph{Nodes.} $V = \mathcal{O} \cup X(s) \cup g$, where:
\begin{itemize}
    \item $\mathcal{O}$: object nodes (one per domain object)
    \item $X(s) = X_p(s) \cup X_n(s)$: state variable nodes (true propositions and numeric fluents)
    \item $g$: goal condition nodes
\end{itemize}

\paragraph{Edges.} For each grounded predicate ${p = \sigma(o_1, \ldots, o_{\text{ar}(\sigma)}) \in X(s) \cup g_p}$:
\begin{equation*}
E \supseteq \{(p, o_i) \mid i \in [1, \text{ar}(\sigma)]\}
\end{equation*}

\paragraph{Node Features.} Categorical features $\mathbf{F}_{\text{cat}}: V \rightarrow \Sigma_V$ encode node semantics:

\begin{equation*}
\mathbf{F}_{\text{cat}}(u) = \begin{cases}
\texttt{object} & \text{if } u \in \mathcal{O} \setminus \mathcal{O}_{\text{const}} \\
u & \text{if } u \in \mathcal{O}_{\text{const}} \\
(\texttt{pred}(u), \texttt{apg}) & \text{if } u \in X_p(s) \cap g_p \\
(\texttt{pred}(u), \texttt{upg}) & \text{if } u \in g_p \setminus X_p(s) \\
(\texttt{pred}(u), \texttt{apn}) & \text{if } u \in X_p(s) \setminus g_p
\end{cases}
\end{equation*}

\paragraph{Edge Labels.} $\mathbf{L}: E \rightarrow \mathbb{N}$ encodes argument position: $\mathbf{L}(p, o_i) = i$.

\subsection{Feature Extraction}

Given the ILG $G_{s,g}$ and $k$ WL iterations, we extract a fixed-dimensional feature vector $\phi(s,g) \in \mathbb{R}^D$ as follows:

\begin{enumerate}
    \item \textbf{Color Refinement}: Run $k$ iterations of WL on $G_{s,g}$, producing color multiset $\mathcal{C}^{(k)}$.
    \item \textbf{Vocabulary Construction}: During training, collect all unique colors across all training graphs to form vocabulary $\mathcal{V} = \{c_1, \ldots, c_D\}$.
    \item \textbf{Histogram Embedding}: For each graph, compute:
    \begin{equation*}
    \phi(s,g)_i = \textsc{Count}(\mathcal{C}^{(k)}, c_i), \quad i \in [1, D]
    \end{equation*}
\end{enumerate}

\subsection{Implementation Details}

We use the \texttt{wlplan} library \cite{chen-wlplan-2024} for WL feature extraction with the following configuration:

\begin{itemize}
    \item \textbf{Graph representation}: Instance Learning Graph (ILG)
    \item \textbf{Iterations}: $k = 2$
    \item \textbf{Hash function}: Multiset hash (deterministic)
    \item \textbf{Pruning}: None
\end{itemize}

\paragraph{Vocabulary Sizes.} The resulting vocabulary sizes (feature dimensions $D$) per domain are:
\begin{center}
\begin{tabular}{lc}
\toprule
\textbf{Domain} & \textbf{Dimension $D$} \\
\midrule
Blocksworld & 587 \\
Gripper & 412 \\
Logistics & 723 \\
VisitAll & 498 \\
\bottomrule
\end{tabular}
\end{center}

\subsection{Properties of WL Embeddings}

\paragraph{Permutation Invariance.} WL embeddings are invariant to object renaming: for any bijection $\sigma: \mathcal{O} \rightarrow \mathcal{O}$, we have $\phi(\sigma(s), \sigma(g)) = \phi(s, g)$.

\paragraph{Size Invariance.} The dimension $D$ depends only on the domain's predicate structure and training distribution, not on $|\mathcal{O}|$. A model trained on 4-block problems produces embeddings of the same dimension for 100-block problems.

\paragraph{Expressivity.} WL embeddings are exactly as expressive as 1-WL message-passing GNNs \cite{xu2018powerful}: two graphs receive the same embedding if and only if 1-WL cannot distinguish them.

\paragraph{Computational Complexity.} For a graph with $n$ nodes, maximum degree $\delta$, and $k$ iterations, the WL algorithm runs in $O(nk\delta)$ time, which is linear in graph size for bounded-degree graphs typical in planning.

\section{Fixed-Size Factored Encoding Details}
\label{sec:fsf_details}

While factored representations have been extensively studied in planning \cite{boutilier2000stochastic,guestrin2003efficient}, our Fixed-Size Factored (FSF) encodings deliberately omit relational structure to serve as a controlled ablation baseline. FSF represents states as vectors of fixed dimension, where each dimension corresponds to a specific object slot with domain-specific semantics. This design isolates the contribution of permutation and size invariance to generalization by providing a representation that: (i) requires a predetermined maximum object count, (ii) depends on fixed object-to-slot mappings, and (iii) necessitates domain-specific manual design.

\subsection{General Structure}

FSF encodings have the form $\phi_{\text{FSF}}(s) \in \mathbb{R}^{N+1}$, where $N$ is the maximum number of objects across all problems in the domain. The encoding consists of:
\begin{itemize}
    \item \textbf{Slot 0}: Global/robot state information
    \item \textbf{Slots 1--$N$}: Per-object state information
\end{itemize}

\paragraph{Special Values.}
\begin{itemize}
    \item $-99.0$: Padding (slot unused for this problem size)
    \item $-10.0$: Don't-care (goal variable not specified)
\end{itemize}

\subsection{Domain-Specific Semantics}

Table~\ref{tab:fsf_detailed} provides detailed semantics for each domain.

\begin{table*}[htbp]
\centering
\caption{Detailed FSF encoding semantics by domain.}
\small
\label{tab:fsf_detailed}
\resizebox{\textwidth}{!}{%
\begin{tabular}{lp{4cm}p{8cm}}
\toprule
\textbf{Domain} & \textbf{Slot Interpretation} & \textbf{Value Semantics} \\
\midrule
\multirow{3}{*}{Blocksworld} 
& Slot 0: Unused (constant 0) & --- \\
& Slot $i$ ($i > 0$): Block $i$ & $0$ = on table; $-1$ = held by gripper; $j > 0$ = on block $j$ \\
\midrule
\multirow{4}{*}{Gripper}
& Slot 0: Robot location & Room index where robot is located \\
& Slot $i$ (ball): Ball $i$ location & Room index if at room; $-j$ if carried by gripper $j$ \\
& Slot $i$ (gripper): Gripper $i$ status & $0$ = free; $j > 0$ = holding ball $j$ \\
\midrule
\multirow{3}{*}{Logistics}
& Slot 0: Unused (constant 0) & --- \\
& Slot $i$: Object $i$ (pkg/truck/plane) & Location index if at location; $-j$ if inside vehicle $j$ \\
\midrule
\multirow{3}{*}{VisitAll}
& Slot 0: Robot position & Cell index where robot is located \\
& Slot $i$ ($i > 0$): Cell $i$ & $0$ = unvisited; $1$ = visited \\
\bottomrule
\end{tabular}%
}
\end{table*}

\subsection{Example: Blocksworld Encoding}

Consider a 4-block Blocksworld state where:
\begin{itemize}
    \item Block A is on the table
    \item Block B is on Block A
    \item Block C is being held
    \item Block D is on the table
\end{itemize}

With object ordering $\{A \mapsto 1, B \mapsto 2, C \mapsto 3, D \mapsto 4\}$:
$$\phi_{\text{FSF}}(s) = [0, \underbrace{0}_{\text{A on table}}, \underbrace{1}_{\text{B on A}}, \underbrace{-1}_{\text{C held}}, \underbrace{0}_{\text{D on table}}, -99, \ldots]$$

\subsection{Limitations of FSF Encodings}

\paragraph{No Size Invariance.}
FSF encodings require a predetermined maximum object count $N$. Problems with $|\mathcal{O}| > N$ cannot be represented. This fundamentally limits extrapolation to larger instances.

\paragraph{No Permutation Invariance.}
FSF encodings depend on a fixed object-to-slot mapping. Different orderings of the same objects yield different vectors, preventing generalization across equivalent states.

\paragraph{Domain-Specific Design.}
Each domain requires manual design of the encoding semantics, limiting applicability to new domains without expert knowledge.

These limitations motivate the use of WL embeddings, which provide permutation and size invariance without domain-specific engineering.

\section{Technical Implementation Pipeline}
\label{sec:impl_pipeline}

This section describes the complete technical pipeline from symbolic planning instances to trained neural transition models, enabling full reproducibility of our experimental results.

\subsection{Data Generation Pipeline}

The data generation process transforms raw PDDL domain and problem files into machine-learning-ready state trajectory embeddings through four sequential stages.

\paragraph{Stage 1: Symbolic Plan Generation.}
We use Fast Downward with a two-tier solving strategy to maximize coverage across problem difficulties. The baseline configuration runs A* search with the landmark-cut admissible heuristic under a 60-second timeout. For problems that exceed this limit, we invoke a fallback configuration using greedy best-first search with the FF heuristic and a 300-second timeout. This staged approach achieves high coverage on training instances (which are intentionally kept small) while maintaining plan quality where possible. Plans are written in standard PDDL action-sequence format.

\paragraph{Stage 2: State Trajectory Reconstruction.}
Given a valid plan, we reconstruct the complete sequence of intermediate world states using the VAL plan validator in verbose mode. VAL applies each action symbolically and outputs the predicates added or deleted at each timestep. We parse this output to build the full trajectory $\langle s_0, s_1, \ldots, s_T \rangle$ where each $s_t$ is represented as a sorted list of ground predicates. This reconstruction is necessary because Fast Downward's search only maintains heuristic state information, not the explicit symbolic states required for supervised learning. The resulting trajectory files are saved in plain-text format with one state per line.

\paragraph{Stage 3: Weisfeiler-Leman Feature Collection.}
WL embeddings require a fixed vocabulary collected from the training distribution. For each domain, we parse all training trajectory files and construct instance learning graphs $G_{s,g}$ for every state-goal pair encountered. We run $k=2$ iterations of color refinement and collect the multiset of final node colors across all training graphs. These color strings are sorted lexicographically and assigned integer indices to form the vocabulary $\mathcal{V}$. The vocabulary size $D = |\mathcal{V}|$ is domain-dependent but fixed once collected, enabling embeddings of arbitrary test-time problem sizes into $\mathbb{R}^D$.

\paragraph{Stage 4: Trajectory Embedding.}
With the vocabulary established, we embed every trajectory across all data splits. For each state $s_t$ in a trajectory, we construct its instance graph $G_{s_t, g}$, perform $k=2$ WL iterations using the fixed vocabulary $\mathcal{V}$, and compute the normalized color histogram as the embedding $\phi(s_t) \in \mathbb{R}^D$. Goals are embedded analogously. The resulting trajectory is a matrix of shape $[T, D]$ where $T$ is the plan length, stored as a NumPy array alongside a separate goal vector of shape $[D]$. This compact representation supports efficient batch loading during training.

\subsection{Transition Model Training}

\paragraph{Dataset Construction for LSTM.}
The LSTM operates on variable-length sequences. For each problem instance, we load the embedded trajectory matrix $[\phi(s_0), \ldots, \phi(s_T)]$ and the goal vector $\phi(g)$. During training, these are collated into padded batches with a custom collate function that tracks the true sequence length of each trajectory. The goal embedding is replicated across all timesteps to form a constant conditioning signal. For delta-mode training, we compute target residuals $\Delta_t = \phi(s_{t+1}) - \phi(s_t)$ on the fly.

\paragraph{Dataset Construction for XGBoost.}
XGBoost requires flattened tabular input. We extract all consecutive state pairs $(s_t, s_{t+1})$ from every trajectory and construct feature vectors by concatenating the current state embedding, the goal embedding, and (for delta mode) the difference vector. Concretely, each training example is a row of shape $[2D]$ (concatenated state and goal) with a target vector of shape $[D]$ (next state or delta). This flattening procedure is applied independently to training and validation splits, yielding dense matrices suitable for gradient boosting.

\paragraph{LSTM Architecture and Training.}
The LSTM model concatenates the $D$-dimensional state and goal embeddings directly into a $2D$-dimensional input vector, which is fed into a two-layer LSTM with 256 hidden units per layer. The LSTM output is passed through a two-layer feedforward head that projects back to the original $D$-dimensional space.
For state-mode training, we minimize cosine embedding loss between predictions and targets, encouraging alignment in direction. For delta-mode training, we minimize mean squared error on the residual vectors.
We train with the Adam optimizer at a learning rate of $10^{-2}$ for 250 epochs, saving the checkpoint with the lowest validation loss as the final model.

\paragraph{XGBoost Architecture and Training.}
XGBoost is configured for multi-output regression using the squared-error objective. We use histogram-based tree construction on GPU with a maximum tree depth of 8 and a learning rate of 0.1. Training proceeds for up to 1000 boosting rounds with early stopping if validation loss does not improve for 10 consecutive rounds. The model learns an ensemble of regression trees that collectively approximate the mapping from $[\phi(s_t), \phi(g)]$ to $\phi(s_{t+1})$ or $\Delta_t$. The best iteration checkpoint is retained based on validation performance.

\paragraph{Loss Function Selection.}
For state-mode training, we adopt cosine embedding loss rather than mean squared error. This choice reflects the sparse, high-dimensional nature of WL embeddings: states differing by a single predicate may have Euclidean distances dominated by uninformative dimensions, whereas cosine similarity emphasizes directional alignment in the feature space. Empirically, we observe that cosine loss enables more stable convergence for LSTM models predicting full state vectors. For delta-mode training, we use mean squared error directly on the residual vectors $\Delta_t$, as these deltas are inherently sparse and low-magnitude, making component-wise regression more appropriate than angular alignment. XGBoost uses squared error in both modes as it does not support cosine objectives natively, though the distance metric used during inference (cosine for state mode, Euclidean for delta mode) remains consistent with the training objective's geometric assumptions.

\subsection{Inference and Plan Decoding}

\paragraph{Symbolic State Maintenance.}
During test-time execution, we maintain the current symbolic state $s_t$ explicitly as a set of ground predicates. This state is initialized to the problem's $s_0$ and updated only through valid operator applications, ensuring that every intermediate state is well-formed under the domain's transition function $\gamma$.

\paragraph{Neural Successor Prediction.}
At each planning step, we embed the current symbolic state and goal to obtain $\phi(s_t)$ and $\phi(g)$. These embeddings are passed through the learned transition model to produce either a direct next-state prediction $\hat{\phi}(s_{t+1})$ or a residual prediction $\Delta_t$ that is added to $\phi(s_t)$. The resulting target vector $\mathbf{v}_t$ represents the model's internal prediction of where the plan should transition next.

\begin{table}[!t]
\centering
\caption{Complete hyperparameter settings for all models.}
\label{tab:hyperparams_full}
\begin{tabular}{lcc}
\toprule
\textbf{Parameter} & \textbf{LSTM} & \textbf{XGBoost} \\
\midrule
\multicolumn{3}{l}{\textit{Architecture}} \\
Hidden dimension & 256 & --- \\
Number of layers & 2 & --- \\
Tree depth & --- & 8 \\
\midrule
\multicolumn{3}{l}{\textit{Training}} \\
Learning rate & $10^{-2}$ & 0.1 \\
Batch size & 32 & --- \\
Max epochs / rounds & 250 & 1000 \\
Early stopping patience & --- & 10 \\
Optimizer & Adam & --- \\
Loss (state mode) & Cosine & MSE \\
Loss (delta mode) & MSE & MSE \\
\midrule
\multicolumn{3}{l}{\textit{Inference}} \\
Beam width & 3 & 3 \\
Distance metric (state) & Cosine & Cosine \\
Distance metric (delta) & Euclidean & Euclidean \\
\bottomrule
\end{tabular}
\end{table}

\paragraph{Symbolic Successor Enumeration.}
Using the ground operator set $\mathcal{A}$ and applicability preconditions, we enumerate all valid symbolic successors $\mathrm{Succ}(s_t) = \{\gamma(s_t, a) \mid a \in \mathcal{A}, \; a \text{ applicable in } s_t\}$. Each candidate successor is embedded using the same WL procedure as during training, yielding a set of embedding vectors $\{\phi(s') \mid s' \in \mathrm{Succ}(s_t)\}$.
To perform PDDL parsing, grounding, and successor generation during inference, we utilize Pyperplan \cite{alkhazraji-et-al-zenodo2020}. While Pyperplan is designed as a lightweight prototyping tool rather than a state-of-the-art search engine, its accessible Python API is well-suited for our neuro-symbolic successor enumeration step.

\paragraph{Nearest Neighbor Decoding.}
We compute the Euclidean distance (for delta mode) or cosine distance (for state mode) between $\mathbf{v}_t$ and each candidate embedding $\phi(s')$. The candidate with the minimum distance is selected as the next symbolic state $s_{t+1}$, and the unique action $a$ satisfying $\gamma(s_t, a) = s_{t+1}$ is appended to the plan. This guarantees that every generated action is applicable and that state evolution respects the symbolic transition semantics.

\paragraph{Termination.}
Planning terminates when the goal condition $g \subseteq s_t$ is satisfied or when the dynamic
horizon $T_{\max} = \max(100,\; 10 \cdot |\mathcal{O}|)$ is reached.
The resulting action sequence is validated externally using VAL to confirm correctness.

\subsection{Dynamic Planning Horizon}
\label{sec:dynamic_horizon}

The inference horizon (maximum number of planning steps before termination) is set
dynamically rather than as a fixed constant.
Let $N = |\mathcal{O}|$ denote the number of objects in the test instance.
The horizon is computed as

\begin{equation*}
    T_{\max} = \max(T_{\text{base}},\; c \cdot N),
\end{equation*}

where $T_{\text{base}} = 100$ is the baseline cap inherited from training and $c = 10$ is a
per-object multiplier.
The motivation is that optimal plan length in many STRIPS domains scales at least linearly with
$|\mathcal{O}|$: in \textit{VisitAll} the goal conjunct grows as $\Theta(N)$, and in
\textit{Gripper} each ball requires at least two transport steps.
A fixed cap of 100 steps therefore becomes a binding constraint precisely on the large
extrapolation instances where the model is asked to generalize.
Replacing it with $10 \cdot N$ avoids premature termination without requiring any
domain-specific tuning.
The factor $c = 10$ was chosen conservatively to exceed the worst-case optimal plan length
observed in the training distribution; we did not tune it beyond this sanity check.

\subsection{Computational Resources}

All experiments were conducted on an HPC cluster with the following specifications:
\begin{itemize}
    \item \textbf{GPU}: NVIDIA H100
    \item \textbf{CPU}: 128 cores per node (for data generation)
    \item \textbf{Memory}: 256GB RAM per node
\end{itemize}

Total compute time: approximately 1 GPU-hour for all experiments.

\subsection{Hyperparameters}

Table~\ref{tab:hyperparams_full} provides complete hyperparameter settings.